\documentclass[letterpaper,10pt,conference]{ieeeconf}
\usepackage[T1]{fontenc}
\usepackage{graphicx}
\usepackage{xcolor}
\usepackage{booktabs}
\usepackage{amsmath}
\makeatletter
\let\NAT@parse\undefined
\makeatother
\usepackage[hidelinks]{hyperref}
\DeclareFontShape{T1}{ptm}{b}{it}{<->ssub * ptm/b/n}{}
\DeclareFontShape{T1}{ptm}{bx}{it}{<->ssub * ptm/b/n}{}
\IEEEoverridecommandlockouts
\title{\LARGE\bf RoboColiseum: Co-evolution of Simulation, Evaluation, and Models}
\author{Chenghao Yin$^{*}$, Da Huang$^{*}$, Di Yang$^{*}$, Jichao Wang$^{*}$, Junhui Wu$^{*}$, Nanshu Zhao$^{*}$,\\
Tong Ran$^{*}$, Zhijun Li$^{*}$, Gang Zhao, Linting Guo, Rui Feng, Sheng Zhang, Tianyu Jiang,\\
Wenjun Sun, Yongxin Su, Zhaobo Liu, Zheyuan Xing, Sukai Wang, Yue Zhou, Qihan Jiang, Jianhua Jiang,\\
Chuang Men, Yuxiang Yan, Daizhen Lv, Kunyu Yu, Yongjian Shen, Tianxiang Zhu, Siying Zhang,\\
Wenhao Wang, Jun Zhao, Bowen Jiang, Zhi Zeng, Maoqing Yao$^{\dagger}$, and Qian Wang$^{\dagger}$%
\thanks{$^{*}$These authors contributed equally. $^{\dagger}$Corresponding authors.}}

\makeatletter
\newenvironment{compactfigure}{\par\medskip\noindent\begin{minipage}{\columnwidth}\def\@captype{figure}}{\end{minipage}\par\medskip}
\newcommand{\frontmatteroverview}{%
\begin{minipage}{\textwidth}
\begin{minipage}{\linewidth}
\begin{abstract}
Simulation-based evaluation of embodied models requires calibrated interactions, diverse tasks, and explicit execution constraints. We present RoboColiseum, a platform combining system identification, composable physics and rendering, and multidimensional benchmarking. Recorded-command replay characterizes robot dynamics and object interactions, while a shared interface connects rigid and deformable simulation to rendering. Across 22 paired real--simulation robot--task configurations, task scores have a reported Pearson correlation of 0.9686. Eight evaluation tracks span instruction following, spatial reasoning, general manipulation, robustness, dexterous manipulation, precision manipulation, whole-body coordination, and execution across robot embodiments. Budget-aware scoring combines task progress, resource use, and completion rewards. We report 12 selected Season~I policies and all 11 systems on the public Season~II leaderboard, comprising eight Agent systems and three VLA policies. Agent2Act connects language models to robot-control execution; the VLA study uses mixed-task supervised fine-tuning. The results reveal substantial variation across tracks within each evaluation category. RoboColiseum connects simulation development with capability-oriented evaluation and provides a foundation for extending robot interactions and model assessment.
\end{abstract}
\end{minipage}
\par\vspace{8pt}
\centering
\includegraphics[width=\textwidth]{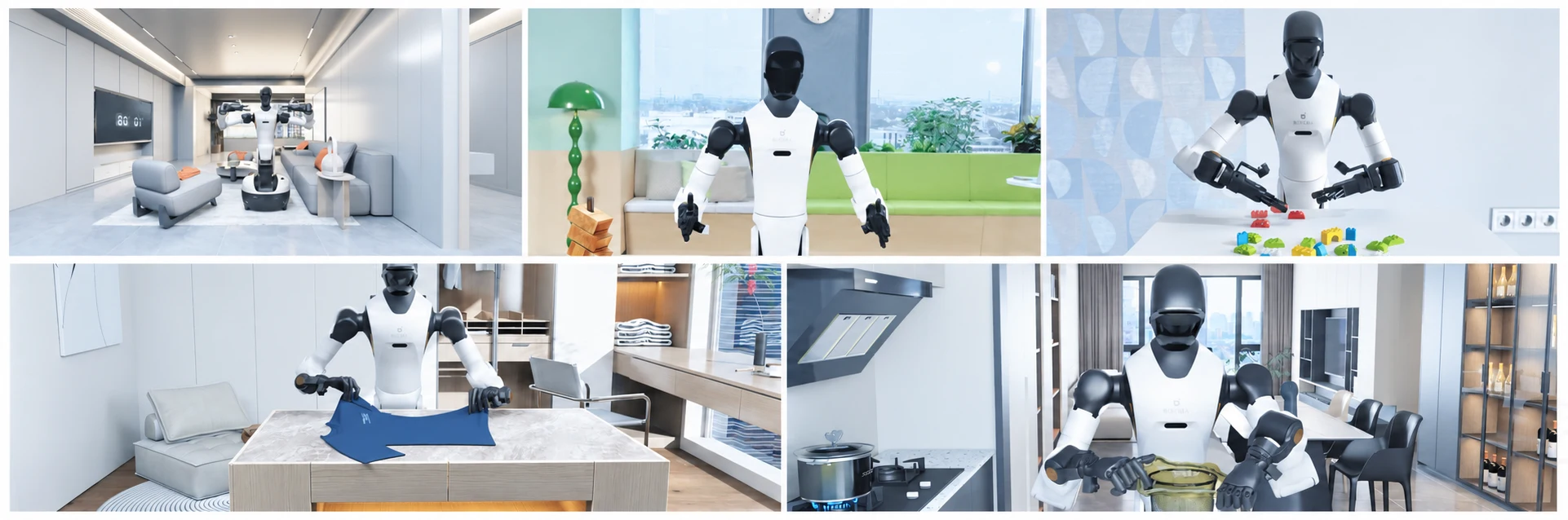}
\def\@captype{figure}
\caption{Examples of robot manipulation in RoboColiseum, including tabletop tasks and cloth handling.}
\label{fig:system-overview}
\par\vspace{10pt}
\end{minipage}%
}
\IEEEaftertitletext{\frontmatteroverview}
\makeatother

\begin{document}
\maketitle
\thispagestyle{empty}
\pagestyle{empty}

\vspace{12pt}
\section{Introduction}
Vision-language-action (VLA) models~\cite{kim2024openvla,black2024pi0} and world action models (WAMs)~\cite{ye2026dreamzero} motivate evaluation that connects instruction understanding with physical execution. Simulation provides repeatable task instances and access to interaction diagnostics, but its usefulness depends on the relation between simulated and real behavior, the interactions it represents, and the information and computational resources available to the evaluated system.

Three challenges motivate our design. First, physical plausibility of identified parameters does not by itself establish accurate task behavior. Physically consistent inertial identification~\cite{wensing2018physicalconsistency} and object parameter estimation~\cite{he2026rigpi} provide tools for calibration, while paired real--simulation evaluation is needed to assess the resulting task scores. Second, rigid and deformable interactions require different numerical representations and contact models. SoftMAC~\cite{liu2024softmac} and Embedded IPC~\cite{du2024embeddedipc} illustrate approaches to coupled interaction and computational efficiency. A common interface can make different backends accessible without assuming identical physical behavior or support for every backend combination. Third, evaluation must distinguish reasoning, execution, and resource use. VLABench~\cite{zhang2025vlabench}, EmbodiedBench~\cite{yang2025embodiedbench}, and EmbodiedEval~\cite{cheng2025embodiedeval} motivate broader interactive evaluation. Agent2Act~\cite{robocoliseum2026agent2act} evaluates multimodal models by having them directly generate robot-control commands through shared observation, execution, and feedback interfaces. The evaluated agent includes both the model and its execution harness.

RoboColiseum focuses on three components. System identification and recorded-command replay characterize robot response and object interaction under stated control and observation settings. A shared protocol separates physics from rendering and supports backend-specific rigid and deformable representations. An eight-track benchmark retains instruction following, spatial reasoning, general manipulation, and robustness as Season~I, and adds dexterous manipulation, whole-body coordination, precision manipulation, and cross-embodiment transfer as Season~II.

We assess RoboColiseum through paired Real2Sim experiments and public benchmark results. Season~I retains four foundational tracks~\cite{yin2026seasoni}; Season~II introduces four additional capability dimensions. We report separate results for eight Agent systems and three VLA policies, preserving their respective training and execution settings. Agent2Act~\cite{robocoliseum2026agent2act} connects language-model reasoning to robot-control execution. The evaluation protocol specifies hidden layouts, execution budgets, and submission review. These components support a development cycle in which task-level observations motivate changes to simulation, training data, and evaluation coverage.

\section{System Identification and Real2Sim}
\label{sec:real2sim}
Our objective is to reproduce command-conditioned robot and object responses, rather than only their visual appearance. We separate the robot dynamics, object interaction parameters, and execution settings so that each can be evaluated under a stated configuration. Benchmark task correspondence is evaluated separately in Section~\ref{sec:real2sim-eval}.

\subsection{Real2Sim}
\textbf{State, commands, and observations.} Let $x_k$ contain robot and object configurations and velocities, and let $u_k$ denote recorded joint and gripper targets. A compact representation of replay is
\begin{equation}
\begin{array}{l}
x_{k+1}=f_{\mathcal E}(x_k,u_k;\theta_r,\theta_o,\psi),\\
\widehat y_k=h_{\eta}(x_k).
\end{array}
\label{eq:replay-model}
\end{equation}
Here $\mathcal E$ denotes the physics backend, $\theta_r$ robot parameters, $\theta_o$ object parameters, $\psi$ control and integration settings, and $\eta$ the observation calibration. This notation separates dynamics from observation: image points and RGB-D feature estimates are not direct measurements of the complete physical state.

Each comparison records the embodiment, end effector, scene, initial state, and command interface. In G2 object replay, recorded position targets are held at approximately 100 Hz while physics advances in 1 ms substeps. Object poses are never overwritten and grasps use contact rather than attachments. Thus tracking an object requires both appropriate robot response and an adequate contact model. The gripper-free robot study and G2--Swiftpicker object study retain separately frozen configurations; fixing a configuration does not eliminate its modeling or calibration error.
\begin{compactfigure}
\centering
\includegraphics[width=\columnwidth]{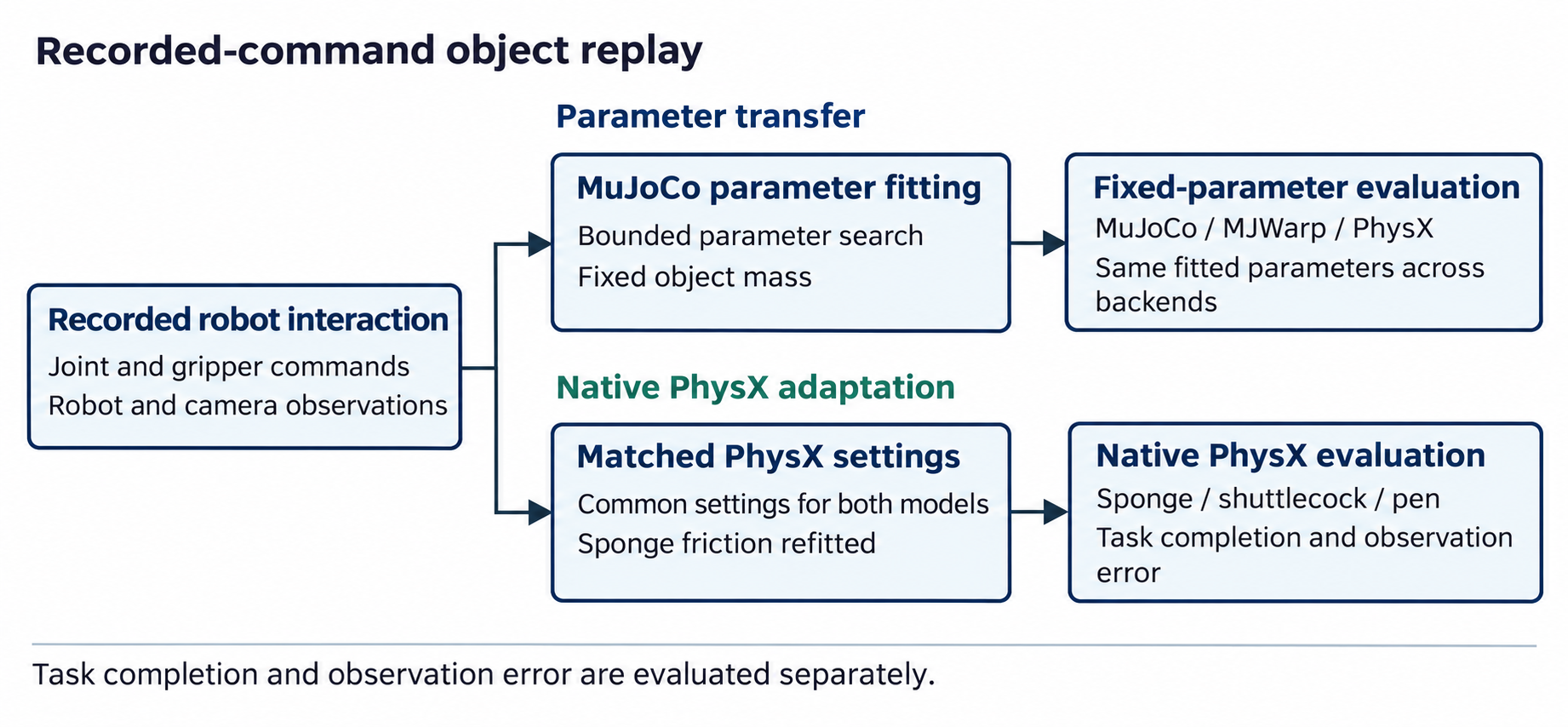}
\caption{Recorded-command object replay. Fixed-parameter transfer evaluates the same fitted object parameters across backends. Native PhysX evaluation uses matched settings and includes a separate sponge-friction refit.}
\label{fig:command-replay}
\end{compactfigure}

\subsection{System Identification}
\textbf{Robot model.} For each seven-joint arm of the G2 crsB platform, we use the following effort balance:
\begin{equation}
\begin{array}{l}
KI=\tau_{\rm rigid}+F\phi(\dot q)+D\dot q+A\ddot q+b,\\
\tau_{\rm rigid}=M(q;\pi)\ddot q+c(q,\dot q;\pi)+g(q;\pi).
\end{array}
\label{eq:robot-model}
\end{equation}
The fixed scale $K$ maps motor current $I$ to actuator-side effort. The inertial parameters $\pi$ determine rigid-body dynamics; diagonal $F,D,A$ describe Coulomb-like friction, viscous damping, and effective armature, while $b$ is an offset. The identification basis is elementwise $\phi(v)=\tanh(v/\epsilon)$, with $\epsilon=0.001$ rad/s. Effective armature may absorb model approximations and is not certified rotor inertia.

\textbf{Staged excitation.} Constant-speed sweeps emphasize friction and damping with little acceleration. Sine motion introduces acceleration-dependent response and estimates armature with temporary inertial parameters. Coupled Fourier motion then estimates inertia and offsets with $F,D,A$ fixed. This ordering limits compensation between friction and inertia; physical-consistency constraints restrict the inertial fit but do not guarantee unique recovery of individual link properties. The effort reconstruction residual is
\begin{equation}
r_k^I=KI_k-\tau_{\rm model}(q_k,\dot q_k,\ddot q_k).
\label{eq:effort-residual}
\end{equation}
The ARX X5 model uses a simplified identification setting because force control is unavailable in that study.

\textbf{Object model and objective.} Under fixed robot execution, bounded candidate searches adjust effective contact parameters and, where applicable, geometry, center of mass, and initial orientation; object mass priors remain fixed. The five assets cover carrying, effective compliant contact, and handover. For valid observation timestamps $\mathcal V$, define
\begin{equation}
e_p=\sqrt{\frac{1}{|\mathcal V|}\sum_{k\in\mathcal V}
\|h_{\eta,p}(x_k)-y_{p,k}\|_2^2}.
\label{eq:object-error}
\end{equation}
The observation map returns a 3D feature in metres or a projected image feature in pixels; errors in these spaces are not pooled. For gum-box and sponge source fitting, the reported objective is
\begin{equation}
J_{\rm obj}=\left(\frac{e_{p,\rm train}}{0.01\,\mathrm{m}}\right)^2
+0.1\left(\frac{e_{g,\rm hold}}{0.03\,\mathrm{rad}}\right)^2,
\label{eq:object-objective}
\end{equation}
where $e_{g,\rm hold}$ is holding-phase gripper RMSE. The denominators normalize units; they are not measured noise levels. Handover fitting uses a different RGB/gripper objective. Task-feasibility gates supplement observation error; a small gripper error alone does not establish a successful grasp.

Freezing execution reduces, but does not remove, ambiguity between object properties and robot, geometry, initial-state, or camera error. Consequently, fitted contact and damping quantities are treated as effective simulation parameters. Candidate searches are bounded by the configured parameter ranges and task-feasibility checks.
\begin{compactfigure}
\centering
{\scriptsize\textcolor[HTML]{2B628A}{$\times$} Baseline RGB error\quad\textcolor[HTML]{397F6F}{$\bullet$} Tested-model RGB error}\par\smallskip
\includegraphics[width=\columnwidth]{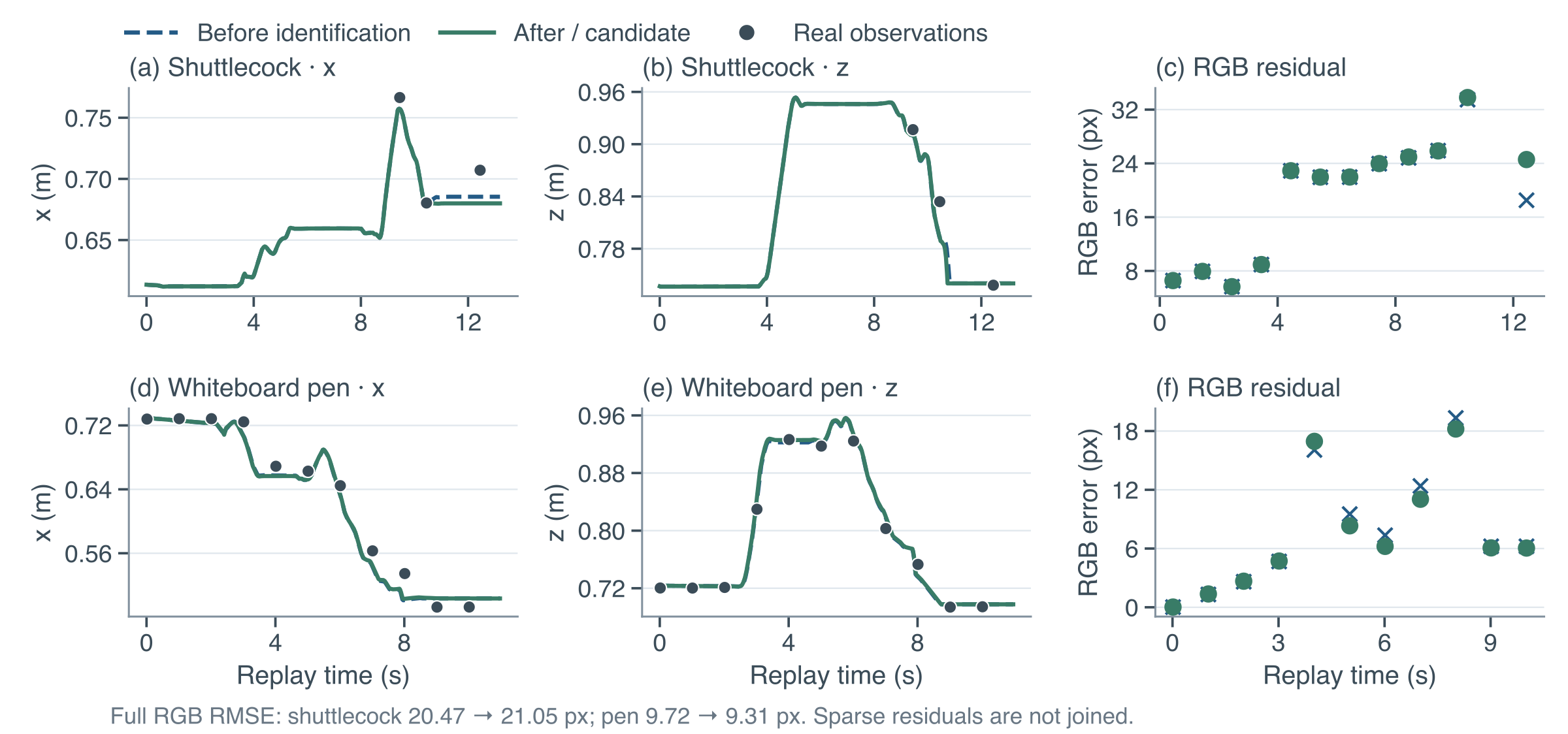}
\caption{Native PhysX handover diagnostics for shuttlecock (top) and pen (bottom). In the position panels, dashed blue and solid teal denote baseline and tested-model trajectories; dark points are real observations. In the RGB-residual panels, blue crosses and teal circles denote baseline and tested-model errors at valid observation times. The printed RMSE values summarize the full valid-time RGB observations, rather than only the plotted examples.}
\label{fig:object-protocol}
\end{compactfigure}

\subsection{Simulation Calibration and Validation}
\textbf{Controller and sensor model.} The seven-joint arm in the gripper-free G2 study uses position PD control; before configured actuator limits, its drive and sensor readout are represented as
\begin{equation}
\begin{array}{l}
\tau_{\rm drive}=K_p(q_{\rm cmd}-q_{\rm sim})-K_d\dot q_{\rm sim},\\
\widehat\tau_{\rm sensor}=\tau_{\rm link,sim}+f_{\rm sensor}(\dot q_{\rm sim},h).
\end{array}
\label{eq:control-readout}
\end{equation}
The adopted $K_p=1000\mathbf{I}_7$ and $K_d=\mathrm{diag}(1,5,1,1,1,1,1)$ use units N\,m/rad and N\,m\,s/rad, respectively; these are simulation settings, not identified hardware gains. Here $\mathbf{I}_7$ is the identity matrix and $h$ denotes observer history. The sensor-side correction is postprocessing only: it changes neither plant forces nor control. Current-derived effort in Eq.~\ref{eq:robot-model}, link-side simulated torque, and measured joint-side sensor feedback are distinct quantities. Physical friction opposes motion in forward dynamics; the inverse-dynamics identification basis and forward-replay friction implementation are documented separately.

\textbf{Validation hierarchy.} Inverse-dynamics reconstruction tests whether the model explains recorded effort; forward prediction tests its motion response; closed-loop replay tests command-conditioned behavior under the adopted controller. Strong feedback can suppress position error despite an imperfect plant, so position and uncentred sensor-torque RMSE are reported separately (Fig.~\ref{fig:robot-validation}).

\begin{figure*}[t]
\centering
\includegraphics[width=\textwidth]{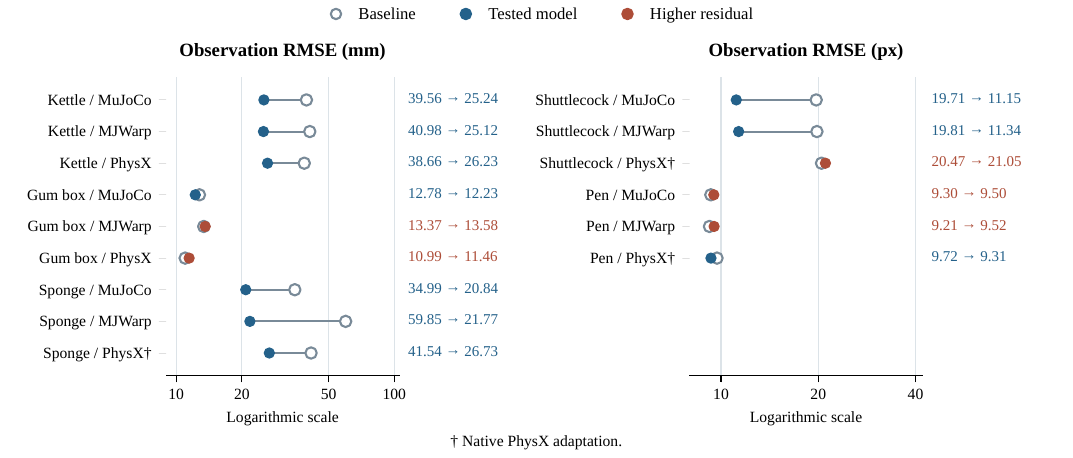}
\caption{Object-observation RMSE for baseline and tested parameter configurations, reported separately in millimetres and pixels. Open markers denote the baseline; filled markers denote the tested model, with red indicating a larger residual. $\dagger$ identifies native PhysX adaptation, including the sponge refit. Pen results describe a parameter candidate that was not adopted.}
\label{fig:object-validation}
\end{figure*}

For objects, fixed-parameter transfer and native engine adaptation answer different questions. The former freezes a candidate across stacks; the latter compares baseline and candidate under matched native settings, with any refit identified separately. Observation error and task completion remain separate readouts (Fig.~\ref{fig:object-validation}). MuJoCo/MJWarp share a contact family, while same-recording temporal checks are not independent-episode validation. Detailed outcomes and failure cases are reported in Section~\ref{sec:real2sim-eval}.
\begin{compactfigure}
\centering
\includegraphics[width=\columnwidth]{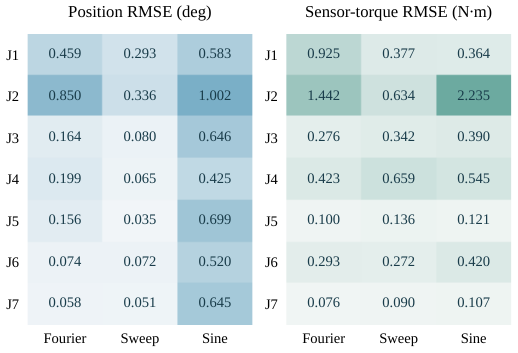}
\caption{Joint-wise replay RMSE for seven joints under Fourier, constant-speed, and sine excitations. Position error is reported in degrees and uncentred sensor-torque error in N\,m. Each column summarizes its corresponding excitation record; Sweep denotes constant-speed motion. Darker cells indicate larger errors within each panel.}
\label{fig:robot-validation}
\end{compactfigure}

\vspace{12pt}
\section{Composable Physics and Rendering}
\label{sec:composable-engine}
Robotic simulation combines interaction models with visual observations whose requirements vary across tasks. RoboColiseum separates physical simulation, rendering, and application logic through a common interface. Physics and rendering components can therefore be composed according to the robot, material, and observation requirements, while retaining a shared control and state-exchange boundary for rigid and deformable manipulation.

\vspace{6pt}
\subsection{Decoupled Architecture and RoboColiseum Protocol}
The \textbf{RoboColiseum Protocol} defines the interfaces and data exchanged among control, physics, and rendering components (Fig.~\ref{fig:engine-composition}). It covers control commands, joint states, rigid-body poses, simulation time, and the geometry updates required to represent deformation. Physics owns the simulated state and publishes its evolution; rendering consumes that state to generate observations. Scene metadata associates physical entities with their visual representations, while simulation timestamps identify the physical states underlying the observations.

Isaac Sim and OVRTX provide alternative rendering paths over this common boundary. Their camera and appearance settings can be configured independently of the selected physical model. Separate-process execution isolates physics and rendering dependencies and permits different update rates. For deformable objects, topology information and evolving vertex positions supplement rigid transforms, allowing shape changes to propagate to the visual scene. The interface separates state exchange from backend-specific numerical methods; available implementations determine which physics--rendering configurations can be used.

\begin{compactfigure}
\centering
\includegraphics[width=\columnwidth]{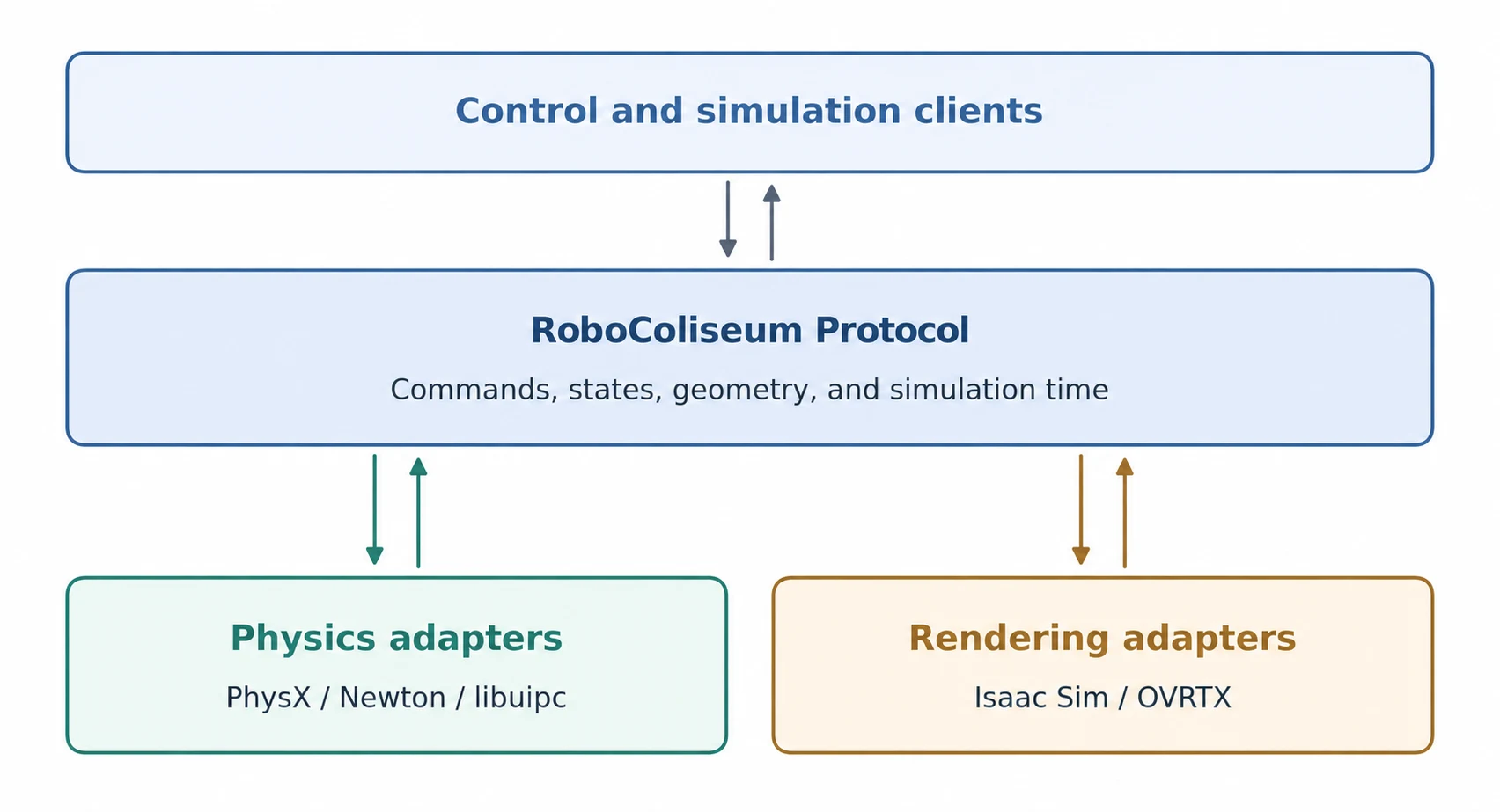}
\caption{Physics and rendering adapters exchange commands, states, geometry, and simulation time through the RoboColiseum Protocol. MuJoCo also supplies solver integration and the auxiliary robot-dynamics worker used by IPC; it is not a separate adapter in this diagram.}
\label{fig:engine-composition}
\end{compactfigure}

\vspace{6pt}
\subsection{Pluggable Physics Engines}
Physics adapters implement a common contract for initialization, command application, stepping, state readback, and reset. Shared execution components coordinate these operations without depending on a particular solver implementation. Scene assembly prepares robot and scene assets together with a manifest that identifies the resources consumed by physics and rendering. Backend-specific conversion stages accommodate differences in asset representation while preserving the common application interface.

The architecture integrates PhysX, Newton, and libuipc backends. PhysX provides an Isaac Sim physics path, Newton supports standalone and Isaac-integrated execution, and libuipc supplies a dedicated rigid--deformable contact backend. MuJoCo participates through solver integration and a cooperating CPU robot-dynamics worker. This distinction separates the engine adapter from the numerical solver or auxiliary dynamics component it uses. New engines can register an adapter, implement the execution contract, and provide the necessary assembly stages. Each backend retains its material models and solver settings, allowing extensions without imposing a single numerical formulation on all tasks.

\vspace{6pt}
\subsection{Rigid--Deformable Interaction}
The libuipc backend extends the engine to deformable manipulation through Incremental Potential Contact (IPC). Cloth and bag-like objects use triangular shell models for membrane deformation and bending. Paper additionally supports plastic bending to retain creases after loading. Volumetric soft bodies use tetrahedral finite elements with a stable Neo-Hookean constitutive model, while rigid objects use affine-body dynamics. These representations place rigid and deformable task objects within the native libuipc contact system.

Contact, friction, and self-contact connect object deformation to robot motion. Material properties, shell thickness, and collision geometry are specified by the scene, enabling different interaction models within the same execution framework. Geometry mappings transfer solved positions to the authored visual meshes so that rendering follows the physical deformation. The supplied scene configurations cover cloth folding, paper folding, cloth-hanger manipulation, and drawstring-bag manipulation, illustrating how the same backend represents different deformable interaction structures.

\vspace{6pt}
\subsection{Robot Dynamics and IPC Coordination}
Deformable manipulation combines articulated robot dynamics with nonlinear contact and material deformation. RoboColiseum assigns these computations to cooperating components. A CPU MuJoCo worker advances robot dynamics and configured robot--static-environment contacts. Robot link poses and velocities then provide kinematic inputs to IPC, which advances task objects and deformable contact. The two components operate at different rates, with the IPC driver selecting the latest valid robot state for each native step.

This decomposition is intended to reduce the robot-dynamics burden within the IPC solve and concentrate its iterations on deformable interaction. The libuipc backend owns native stepping and dynamic task state, while Newton SDK utilities provide model construction, state arrays, and forward kinematics. Commanded forward kinematics or external joint measurements can also supply robot motion through the same input boundary. The current coupling is directional: robot motion drives IPC without feeding IPC contact forces back into the CPU robot model. Completed rigid and deformable states are subsequently exposed through the shared interfaces for rendering and data collection.

\section{Benchmark Design and Evaluation Protocol}
\label{sec:benchmark}

\subsection{Season I: Foundational Tracks}
Season~I organizes foundational interactive capabilities into \emph{Instruction Following}, \emph{Spatial Reasoning}, \emph{General Manipulation}, and \emph{Robustness}. Tasks connect language and visual observations to physical actions while emphasizing target selection, spatial relationships, action sequences, or sensitivity to evaluation conditions. Published task-design principles combine motor primitives, cognitive requirements, and task complexity~\cite{yin2026seasoni}. Longer interactions compose primitives and increase planning horizon or coordination demands.

\textbf{Instruction Following.} The published instruction suite contains ten tasks. Object-selection tasks vary the language needed to identify a target among distractors. Instructions refer to color, shape, size, numerical identity, object category, common-sense needs, or logical alternatives. Examples include selecting a colored block or an object appropriate to a stated need. Retaining a pickup interaction while changing the requested attribute emphasizes language grounding and target disambiguation.

\textbf{Spatial Reasoning.} Eight spatial tasks require interpreting absolute locations and relations between objects, then using those relations to select or construct a target arrangement. Examples include selecting an object by relative distance, placing an object to the left of another, ordering blocks by size, and stacking blocks. These interactions examine scene interpretation, relational target specification, and the geometric organization of successive actions.

\textbf{General Manipulation.} Ten manipulation tasks extend pickup to coordinated sequences of grasping, transport, placement, and object-state changes. Examples include moving a pot with both hands, clearing a desktop, pouring workpieces, scooping popcorn, and sorting packages. The tasks emphasize operation-specific grasp selection, object control during motion, and ordering dependent actions to complete an interaction.

\textbf{Robustness.} Five perturbation categories retain the underlying goal while changing background appearance and illumination, camera pose or image quality, robot initial pose, or instruction phrasing. Configured image degradations include noise, occlusion, and dropped frames. These variations probe the stability of target selection and execution across observation and initialization conditions. Organizing variants by perturbation type supports analysis of which changes are associated with performance sensitivity.

Together, the four tracks relate instruction interpretation, geometric reasoning, action sequencing, and stability across conditions to observable task behavior.

\subsection{Season II: Task Design}
As model capabilities and research objectives change, additional tasks can expose limitations that established tracks no longer distinguish. Season~II extends the foundational tracks with \emph{Dexterous Manipulation}, \emph{Whole-Body Coordination}, \emph{Precision Manipulation}, and \emph{Cross-Embodiment Transfer}. These additions introduce functional finger interactions, coupled mobility and manipulation, fine alignment of small objects, and action adaptation across robot structures.

\begin{table*}[t]
\centering
\caption{Season~II task families and their intended capability emphasis. Cross-embodiment transfer denotes evaluation across robot configurations; it does not imply a held-out-embodiment training protocol.}
\label{tab:benchmark-dimensions}
\small
\renewcommand{\arraystretch}{1.15}
\begin{tabular}{@{}p{0.20\textwidth}p{0.34\textwidth}p{0.41\textwidth}@{}}
\toprule
\textbf{Dimension} & \textbf{Representative tasks} & \textbf{Capability emphasis} \\
\midrule
Dexterous manipulation & Book retrieval, scissors operation, parcel transport, cup lidding, bowl stacking; also cloth, bag, cube, and pan tasks & Functional finger configurations, coordinated object interaction, and bimanual handling \\
Whole-body coordination & Tray delivery, trash collection, door interaction, shelf placement & Coordinating the mobile base, torso, and arms across navigation and manipulation \\
Precision manipulation & Aperture insertion, connector mating, assembly, packaging & Local geometry perception, fine relative-pose control, and sequential assembly \\
Cross-embodiment transfer & Six shared tasks on four robot configurations & Adapting actions to robot structure and workspace while preserving task intent \\
\bottomrule
\end{tabular}
\end{table*}

Task design starts from the capability needed to complete an interaction. Within each added suite, tasks share a central capability requirement while varying the objects, geometry, and action sequence. Table~\ref{tab:benchmark-dimensions} summarizes the four added suites. The following subsections describe the four Season~II suites and their intended capability requirements.

The benchmark accommodates VLA policies, WAM policies, and model-driven agent systems. VLA models generate actions from visual observations and language instructions, optionally conditioned on robot state~\cite{kim2024openvla,black2024pi0}. WAMs such as DreamZero jointly model visual dynamics and actions~\cite{ye2026dreamzero}. Agent systems combine model-driven reasoning with observations, planning, tool use, and execution feedback. In Agent2Act~\cite{robocoliseum2026agent2act}, models directly generate robot-control commands through shared observation, action-execution, and feedback interfaces. The model and its execution framework together form the evaluated system.

Together, the eight dimensions connect task demands to track-level performance profiles. Their results support capability-specific selection of demonstration data and training tasks. Comparisons across tracks describe performance patterns; isolating individual failure mechanisms requires controlled experiments.

\subsubsection{Dexterous Manipulation}
This suite uses the G2 platform equipped with Wuji dexterous hands to examine how multiple fingers participate in object interaction. Book retrieval, scissors operation, bimanual parcel transport, cup lidding, and bowl stacking require different combinations of finger configuration, manipulation after grasping, and hand coordination. These tasks extend evaluation to the functional role of hand motion within an interaction.

Scissors operation makes this requirement explicit. A thumb and another finger engage separate handle holes and coordinate their motion to open and close the scissors. A conventional parallel-jaw gripper cannot reproduce this finger-in-hole operating pattern. The task examines whether a model can infer how to use the tool, form an appropriate hand configuration, and drive its motion while maintaining the hold. Book retrieval instead begins with a constrained workspace: the robot tilts a book within a tightly arranged shelf before extracting it, linking contact selection to the sequence needed to create removal space.

The remaining tasks broaden the coordination demands. Parcel transport requires both hands to control one object, while cup lidding and bowl stacking connect grasp configuration to subsequent approach, alignment, and placement. Comparing these interactions helps identify relative weaknesses in functional grasp planning, sustained manipulation, and bimanual coordination.

The evaluated VLA suite contains ten dexterous task configurations: the five interactions above, together with cloth hanging, drawstring-bag manipulation, cloth folding, Rubik's Cube solving, and pan tossing. These additions introduce deformable handling and articulated or dynamic object manipulation.

\subsubsection{Whole-Body Coordination}
Whole-body tasks extend evaluation to interactions that require the mobile base, torso, and arms to work together. Base position and body posture determine the reachable workspace, while a held object constrains subsequent movement. Tray delivery, trash collection, door interaction, and shelf placement are designed around this coupling over extended action sequences.

Tray delivery connects pickup and transport to maintaining the load during motion. Trash collection links retrieval from different locations to eventual disposal, requiring the model to track progress and choose suitable approach positions. Door interaction couples base placement, articulated-object manipulation, and the order of passage and closing. Shelf placement requires body configuration to adapt to the destination while connecting transport to final placement.

These tasks examine the joint decisions of where to move, how to position the body for interaction, and how to connect the current operation to later actions. They help reveal whether local manipulation skills extend to complete mobile sequences, whether object control is maintained during transport, and whether approach, interaction, and departure form a coherent plan.

The current hidden-set manifest instantiates eight whole-body task configurations: board wiping, conveyor pickup and shelf placement, door passage and closing, roller-shutter entry, trash collection, plate placement in a dish rack, tray-based tissue delivery, and carton transport and sorting. The examples above describe their coordination requirements; the manifest defines the evaluated coverage.

\subsubsection{Precision Manipulation}
Precision tasks focus on interactions in which local geometry and small relative motions affect completion. The suite covers eight task categories: inserting a straw into a milk-tea cup, capping a pen, depositing a coin, inserting a block, connecting an Ethernet cable, assembling a part on a shaft, packaging a phone, and connecting a phone charging cable. Apertures, ports, and mating parts require task understanding to translate into suitable approach directions, alignment, and action transitions.

The tasks cover three recurring demands. Straw insertion, coin deposition, and pen capping require perception of a small object relative to its target and selection of an appropriate approach. Connector mating and shaft assembly emphasize orientation and alignment between parts; phone charging additionally requires the two arms to handle the phone and connector separately. Phone packaging embeds precise placement within a sequence involving the phone, internal packaging components, and box lid.

Varying mating geometry, manipulated objects, and action sequences examines fine-grained spatial perception, relative-pose control, bimanual alignment, and sequential assembly. Task comparisons help identify where execution becomes difficult: locating the target interface, converting an approach into fine alignment, or maintaining established part relationships through subsequent operations.

\subsubsection{Cross-Embodiment Transfer}
The cross-embodiment suite places shared task intentions on ARX, AgileX, G1, and G2 Omnipicker configurations. Six shared tasks cover opening a trash bin and disposing of trash, scraping trash into a bin with a tool, hanging a cup, replacing a keycap, writing, and pouring water. They examine how models adapt articulated-object interaction, tool use, placement, fine manipulation, writing, and liquid transfer to different robot structures.

Differences in kinematic structure, reachable workspace, and end-effector geometry require different action realizations for a shared goal. Opening a trash bin and disposing of trash connects lid manipulation to subsequent placement. Tool-assisted scraping requires robot motion to produce the intended contact between the tool and trash. Cup hanging and keycap replacement require approaches and alignment compatible with the end effector and target geometry. Writing and pouring water connect motion paths and object orientation to the intended interaction. Task instances adapt to each embodiment's workspace while retaining the common semantic objective.

This design adds an embodiment axis to capability profiles. Shared task categories provide a basis for comparing which operations remain stable across robot configurations and which are sensitive to changes in reach, geometry, or control. They also support extending the suite to additional embodiments while preserving the task intent. Performance is interpreted in relation to the training data and adaptation available for each robot. The pooled multi-embodiment fine-tuning used in the present policy study measures execution across included configurations; evaluation on a robot withheld from training would require a separate protocol.

\subsection{Task Authoring, Initialization, and Controlled Variation}
\label{sec:task-instances}
The platform's task-authoring workflow associates a language instruction with robot, object, and environment assets, spatial layouts, initial states, and task constraints. Agent-assisted asset composition and scripted orchestration support variants with different asset combinations, placements, and initial conditions. Episode records retain the task description, scene assets, execution configuration, robot and object states, timestamps, and task outcome. Retaining robot configuration, joint ordering, and camera parameters connects the semantic task to its physical and observational realization. These records support tracing differences between task variants and examining failed executions.

The parallel-simulation path supports reproducible task initialization. Selected environments can restore recorded states and switch between compatible task instances at reset. Scene pre-baking caches settled object poses and support relationships, and layout perturbations use these relationships to keep objects on their supports and clear of one another. Per-environment seeds determine initial-state perturbations independently of how other episodes are scheduled. This task-management mechanism supports replay and controlled comparisons; it is separate from the scoring and inclusion conventions of the historical evaluations reported below.

Controlled variation is organized into scene layout, robot, environment, and rendering factors. These cover object placement; initial arm configurations, base poses, and joint dynamics; lighting, materials, and backgrounds; and camera viewpoints, noise, occlusion, and frame loss. Independent random streams permit selected factors to vary while preserving unrelated random draws. Task-level profiles specify the variation settings, and readback distinguishes requested settings, resolved parameters, and applied changes. This organization makes the perturbation configuration inspectable and supports experiments that separate sensitivity to individual factors from performance across a combined variation profile.

\subsection{Leaderboard Protocol and Evaluation Integrity}
\label{sec:protocol}
The benchmark protocol maintains a dedicated Agent leaderboard separately from the VLA/WAM leaderboard. These categories reflect different inference and action-generation procedures; grouping systems separately preserves their training and execution context. The Season~II policy results contain VLA models only; the public Season~I policy category also includes WAM submissions.

Public task layouts and test-server layouts follow the same intended distribution but instantiate different scenes. The server protocol specifies a fixed official set for scoring and a dynamic hidden set that changes with each submission. Dynamic-set scores are excluded from Overall and are compared with official-set scores to identify discrepancies that may indicate test-set overfitting. Submissions with excessive discrepancies may be restricted from leaderboard inclusion. Human review supplements these checks by examining privileged-information access and evaluation-specific exploits. Together, these checks address layout memorization and inappropriate information access.

The protocol follows the execution-budget and information-access considerations discussed by OpenRUA~\cite{openrua2026revisiting}. Its layout checks and submission review are design safeguards; the reported model scores are not an experimental assessment of their effectiveness.

\subsection{Scoring and Execution Budgets}
\label{sec:scoring}
Task-specific checkers measure intermediate progress and end-to-end completion~\cite{robocoliseum2026protocol}. Execution limits are derived from demonstration data and stored with the task weights in each evaluation configuration. Let $r_{e,j}\in[0,1]$ denote episode $e$'s native score at checkpoint $j$, with normalized task weights $w_j\geq0$ and $\sum_jw_j=1$. For tasks using budget-adjusted policy scoring, the normalized episode score is
\begin{equation}
q_e=\frac{P\sum_jw_jr_{e,j}d_{e,j}+Q C_e}{P+Q}.
\label{eq:episode-score}
\end{equation}
The recorded policy profile assigns $P=100$ process points and $Q=25$ completion points. The indicator $C_e$ equals one only when every required checkpoint is complete, the native task-completion check passes, and all scored stages remain within budget. Unfinished checkpoints earn zero progress. Tasks with standalone milestone scores retain their authored checkpoint definitions.

For the evaluated policy implementation, let $n_{e,j}$ be the number of accepted action dispatches and $L_j\geq30$ the configured stage execution limit. The dispatch allowance and score multiplier are
\begin{equation}
\begin{split}
B_j&=\max\!\left\{1,\left\lfloor\frac45\left\lfloor\frac{L_j}{30}\right\rfloor\right\rfloor\right\},\\
d_{e,j}&=\max\!\left\{0,\min\!\left\{1,3-\frac{2n_{e,j}}{B_j}\right\}\right\}.
\end{split}
\label{eq:policy-budget}
\end{equation}
Thus progress credit is unchanged up to $B_j$ dispatches and decreases linearly to zero at $1.5B_j$; the completion bonus requires $n_{e,j}\leq B_j$. Episode-level budgeted tasks use one shared dispatch count and allowance across checkpoints. These expressions specify the recorded policy configuration. Agent inference-call budgets follow their submission protocol and are distinct from policy action-dispatch counts.

For a track with scored episode set $\mathcal A=\mathcal A_{m,t}$, an episode-weighted policy aggregate is
\begin{equation}
s_{m,t}=\frac{100}{|\mathcal A|}\sum_{e\in\mathcal A}q_e.
\label{eq:track-score}
\end{equation}
The episode set contains records admitted by that evaluation's validity rules; an absent record is not an observed zero. The public leaderboard supplies the reported track aggregates for both categories. We reproduce those aggregates and Overall directly~\cite{robocoliseum2026seasoniileaderboard}, with
\begin{equation}
S_m^{\rm Overall}=\frac14\sum_{t\in\mathcal T_{\rm II}}s_{m,t},
\label{eq:overall-score}
\end{equation}
where $\mathcal T_{\rm II}$ comprises precision manipulation, cross-embodiment evaluation, whole-body coordination, and dexterous manipulation. Overall averages the unrounded track aggregates before two-decimal display rounding. These progress-based scores are distinct from binary success rates.

\section{Experiments}

\subsection{System Identification and Real2Sim Evaluation}
\label{sec:real2sim-eval}
\textbf{Robot response and object replay.} Across 15 long-horizon robot runs, position RMSE ranges from $0.035^{\circ}$ to $1.002^{\circ}$ over 21 joint/excitation cells. Figure~\ref{fig:robot-validation} reports position and uncentred sensor-torque errors separately. These closed-loop replay metrics characterize command-conditioned behavior under the adopted controller.

Object evaluations cover a kettle, gum box, sponge, shuttlecock, and pen. Fixed-parameter transfer and native PhysX adaptation are evaluated separately (Fig.~\ref{fig:command-replay}). Kettle and sponge observation residuals decrease in the displayed comparisons, while gum-box residuals increase under MJWarp and PhysX (Fig.~\ref{fig:object-validation}). Native PhysX handover produces a shuttlecock pixel-RMSE increase from 20.47 to 21.05 and a pen decrease from 9.72 to 9.31. The pen parameter candidate was not adopted; its lower pixel residual alone does not establish task feasibility. The shuttlecock has only three valid depth observations. These results motivate evaluating task completion alongside observation agreement.

\begin{figure*}[t]
\centering
\includegraphics[width=\textwidth]{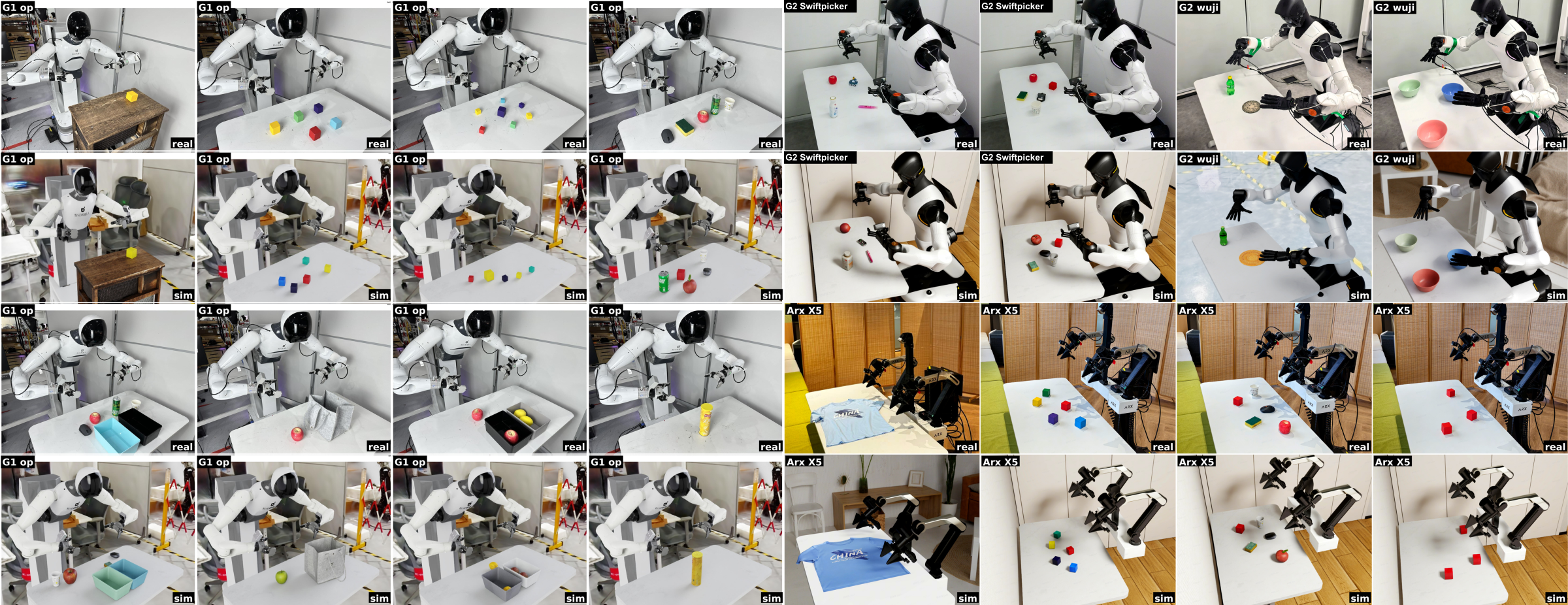}
\caption{Representative real and simulated scenes for foundational manipulation and deformable-object interactions. In each column, real rows 1 and 3 correspond to simulation rows 2 and 4. This montage illustrates the paired scene setting, rather than the coverage of the four Season~II tracks.}
\label{fig:benchmark-scenes}
\end{figure*}

\begin{figure*}[t]
\centering
\includegraphics[width=0.68\textwidth]{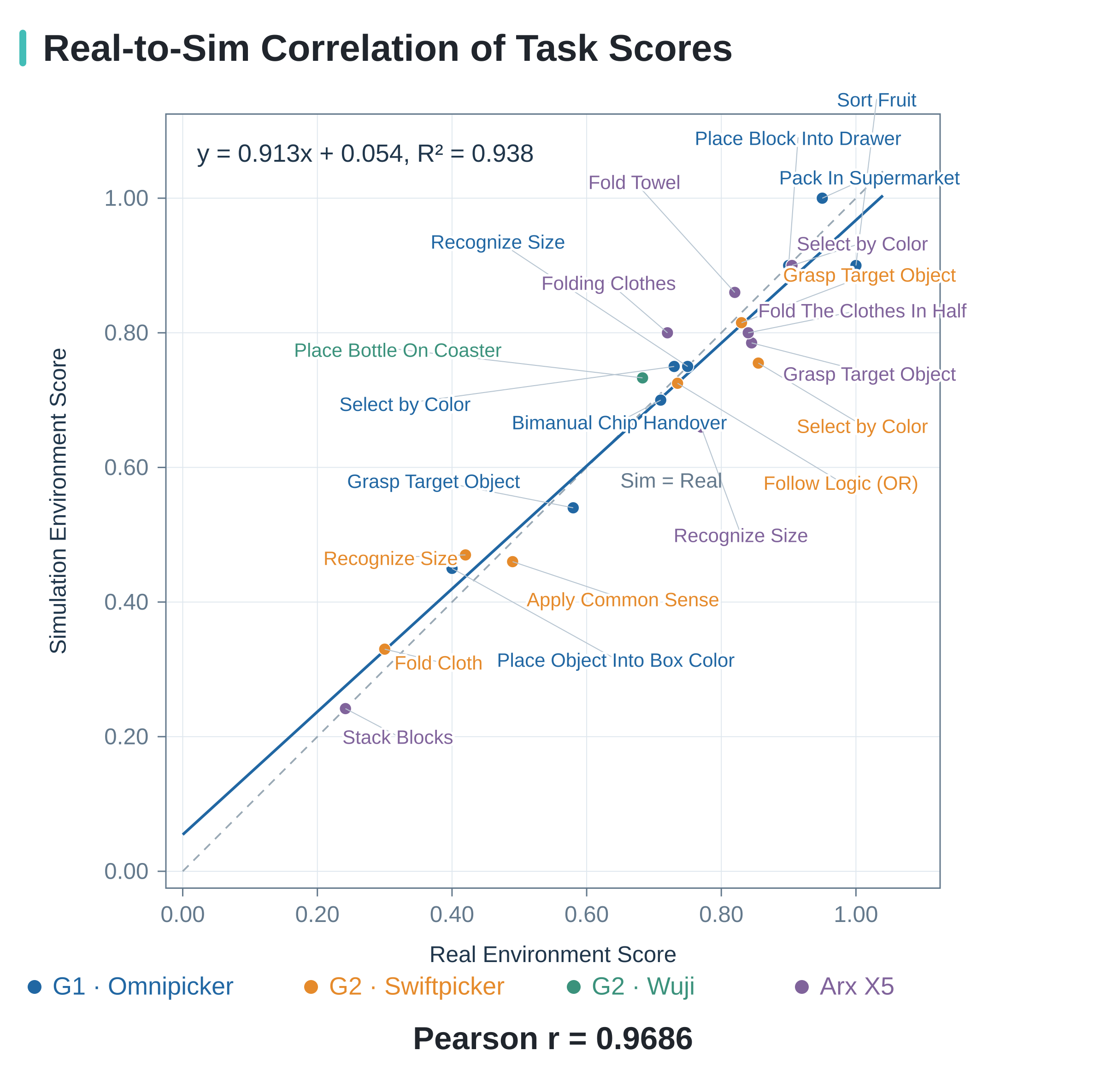}
\caption{Paired real--simulation task scores across 22 robot--task configurations. Colors denote robot configurations; the solid line is the fitted regression and the dashed line denotes equality. Scores are fractional task metrics, separate from the benchmark leaderboard scale.}
\label{fig:benchmark-scores}
\end{figure*}

\textbf{Benchmark-level correspondence.} The paired study covers 22 robot--task configurations spanning 16 task names (Fig.~\ref{fig:benchmark-scenes}). Task scores have reported Pearson $r=0.9686$ across these configurations (Fig.~\ref{fig:benchmark-scores}). Each pair uses its task-specific score definition. The figure summarizes cross-configuration association; it does not estimate the causal effect of parameter identification.

\subsection{Season I: Public Policy Leaderboard}
\label{sec:season-i-results}
Table~\ref{tab:season-i-leaderboard} reports 12 selected policies from the official Season~I leaderboard~\cite{robocoliseum2026seasonileaderboard}. It includes the four baselines discussed by Yin et al.~\cite{yin2026seasoni} and eight additional public submissions. The table uses the official 0--100 scores and the equal-weight four-track Overall.

\begin{table*}[t]
\centering
\caption{Selected Season~I policy results from the official leaderboard (0--100 scale). Rows are sorted by Overall; bold marks the largest value among the listed models.}
\label{tab:season-i-leaderboard}
\small
\renewcommand{\arraystretch}{1.12}
\begin{tabular}{@{}lrrrrr@{}}
\toprule
\textbf{Model} & \textbf{Instruction} & \textbf{Spatial} & \textbf{Manipulation} & \textbf{Robustness} & \textbf{Overall} \\
\midrule
Prime-0 & \textbf{89.61} & 67.91 & \textbf{66.13} & \textbf{86.53} & \textbf{77.54} \\
HiDream-O1-Embodied & 89.13 & \textbf{74.37} & 64.42 & 79.78 & 76.92 \\
DM0.5 & 84.44 & 61.46 & 63.70 & 73.44 & 70.76 \\
GigaBrain-0.7 & 81.66 & 47.29 & 63.20 & 68.00 & 65.04 \\
Xiaomi-Robotics-1 & 76.97 & 34.58 & 60.13 & 65.82 & 59.37 \\
OpenWAM-Alpha & 68.42 & 50.41 & 55.98 & 58.78 & 58.40 \\
$\pi_{0.5}$ & 74.60 & 35.60 & 58.20 & 61.30 & 57.42 \\
ACoT-VLA & 75.70 & 39.70 & 47.70 & 62.20 & 56.32 \\
LingBot-VLA 2.0 & 70.80 & 42.00 & 47.00 & 63.30 & 55.78 \\
GR00T N1.7 & 64.60 & 24.90 & 43.80 & 53.80 & 46.78 \\
InternVLA-A1.5 & 65.00 & 27.40 & 37.00 & 57.50 & 46.73 \\
$\pi_0$ & 36.80 & 13.00 & 34.70 & 31.30 & 28.95 \\
\bottomrule
\end{tabular}
\par\smallskip
\begin{minipage}{0.96\textwidth}\footnotesize Scores are reported on a 0--100 scale. Overall is the equally weighted mean of the four track scores, computed before rounding.\end{minipage}
\end{table*}

Among the listed policies, Prime-0 has the highest Overall (77.54), followed by HiDream-O1-Embodied (76.92) and DM0.5 (70.76). Prime-0 leads instruction following, general manipulation, and robustness within this subset; HiDream-O1-Embodied leads spatial reasoning. These public submissions are interpreted under their respective training and execution configurations.

\subsection{Season II Evaluation Setup}
\label{sec:benchmark-results}
\label{sec:agent-coverage}
The official Season~II leaderboard contains eight Agent systems and three VLA policies~\cite{robocoliseum2026seasoniileaderboard}. Tables~\ref{tab:benchmark-model-comparison} and~\ref{tab:policy-results} retain separate categories for their respective inference and action-generation procedures. Agent systems use Agent2Act~\cite{robocoliseum2026agent2act} to generate robot-control commands from observations and execution feedback.

The VLA evaluation uses mixed-task supervised fine-tuning with 200 demonstrations per task. Each policy has 50 evaluated robot--task configurations and 255 valid episodes: precision has 8 configurations/40 episodes, cross-embodiment 24/120, whole-body 8/45, and dexterous manipulation 10/50. The conveyor task contributes ten episode records; the remaining whole-body configurations contribute five each. The evaluated models include $\pi_{0.5}$~\cite{intelligence2025pi05visionlanguageactionmodelopenworld}, LingBot-VLA-2.0, and GR00T~N1.7. Cross-embodiment results assess the configurations included in mixed-task training.

The tables reproduce all public Season~II entries and their displayed scores. Overall follows Eq.~\ref{eq:overall-score}. Public aggregates characterize the submitted systems; differences in training, task coverage, and resource accounting prevent a controlled comparison of Agent and VLA paradigms.

\begin{table*}[t]
\centering
\caption{Season~II Agent results from the official leaderboard (0--100). Rows follow Overall; bold marks column maxima, except the qualified dexterous value.}
\label{tab:benchmark-model-comparison}
\small
\renewcommand{\arraystretch}{1.12}
\begin{tabular}{@{}lrrrrr@{}}
\toprule
\textbf{Model} & \textbf{Precision} & \textbf{Cross-embodiment} & \textbf{Whole-body} & \textbf{Dexterous} & \textbf{Overall} \\
\midrule
Claude Fable 5.2 & \textbf{33.04} & \textbf{17.99} & 18.42 & 3.73 & \textbf{18.30} \\
GPT-6 Astra & 23.60 & 12.37 & \textbf{18.96} & 0.00 & 13.73 \\
Claude Opus 5.5 & 14.18 & 9.25 & 18.94 & 4.70 & 11.77 \\
Gemini 4 Argon & 11.49 & 8.86 & 17.42 & \textbf{5.19} & 10.74 \\
GPT-6.1 Sol & 15.04 & 8.79 & 17.40 & 0.07 & 10.33 \\
GPT-6.1 Terra & 11.49 & 6.96 & 6.72 & 0.00 & 6.29 \\
DeepSeek-V4.1-Flash & 5.02 & 3.05 & 1.86 & 1.15 & 2.77 \\
Muse Wonder 1.0 & 2.33 & 3.46 & 1.40 & 0.43 & 1.91 \\
\bottomrule
\end{tabular}
\par\smallskip
\begin{minipage}{0.96\textwidth}\footnotesize Overall averages the four unrounded track aggregates.\end{minipage}
\end{table*}

\begin{table*}[t]
\centering
\caption{Season~II VLA policy results from the official leaderboard (0--100). Bold marks the largest value in each column within the policy category.}
\label{tab:policy-results}
\small
\renewcommand{\arraystretch}{1.12}
\begin{tabular}{@{}lrrrrr@{}}
\toprule
\textbf{Model} & \textbf{Precision} & \textbf{Cross-embodiment} & \textbf{Whole-body} & \textbf{Dexterous} & \textbf{Overall} \\
\midrule
$\pi_{0.5}$ & 4.20 & \textbf{10.00} & \textbf{13.87} & \textbf{16.79} & \textbf{11.22} \\
LingBot-VLA-2.0 & 4.40 & 5.70 & 5.33 & 10.66 & 6.52 \\
GR00T N1.7 & \textbf{4.60} & 4.33 & 9.24 & 5.53 & 5.93 \\
\bottomrule
\end{tabular}
\par\smallskip
\begin{minipage}{0.96\textwidth}\footnotesize Scores combine task progress and budget-aware completion. Overall averages four unrounded track aggregates; values reproduce the official two-decimal display.\end{minipage}
\end{table*}

\subsection{Season II Capability Profiles}
Claude Fable~5.2 leads reported Agent Overall (18.30), precision (33.04), and cross-embodiment evaluation (17.99). GPT-6 Astra and Claude Opus~5.5 have similar whole-body scores, 18.96 and 18.94. Without matched repeated evaluations, this 0.02-point separation does not establish a reliable performance difference. Reported Agent dexterous scores remain low, spanning 0.00--5.19; the upper endpoint retains the aggregation qualification in Table~\ref{tab:benchmark-model-comparison}.

Within the policy category, $\pi_{0.5}$ has the highest Overall (11.22) and leads cross-embodiment, whole-body, and dexterous evaluation. GR00T~N1.7 has the highest precision score (4.60), although all three precision scores fall within 4.20--4.60. The policy results identify precision manipulation as a common difficulty under the evaluated training and execution setup. Track-specific profiles motivate focused data collection and controller/model analysis, rather than a single category-wide capability ranking.

\section{Limitations and Future Directions}
\label{sec:limitations}
The Real2Sim evidence covers a limited set of robot--task configurations and observation protocols. Strong feedback can mask plant mismatch, and same-recording replay is not independent-episode validation. Object parameters are effective fits under the adopted geometry, calibration, and controller. Complete calibration records, parameter-search specifications, and paired numerical task scores are needed to assess reproducibility and uncertainty. Directional robot--IPC coupling also omits contact-force feedback to the CPU robot model.

Public benchmark aggregates do not supply full per-case denominators, inference-call counts, or independent-repeat uncertainty. Earlier evaluation records contain heterogeneous invalid-case and unfinished-case rules; reproducing the public rankings does not establish that those rules are matched across systems. Gemini's dexterous sub-suite aggregation remains unresolved. These limitations motivate releasing case-level scores, versioned task weights and budgets, model checkpoints, and training/test splits. The published profiles support within-setting observations; causal comparisons of model families or training procedures require matched experiments. Dynamic-set thresholds and safeguard effectiveness also need quantitative assessment.

The platform's co-evolution principle is to use evaluation findings to guide simulation refinement and training-task selection; quantitative validation of that iterative loop remains future work.

\section{Conclusion}
RoboColiseum integrates system identification, composable physics and rendering, and eight-track embodied evaluation. Paired task scores show strong real--simulation association within the evaluated configurations. Public Season~I and Season~II results provide capability profiles for policy and Agent systems under their respective execution settings. Explicit progress, budget, and completion accounting makes those profiles interpretable. Extending interaction coverage and resolving evaluation-data gaps will support more reproducible simulation and model assessment.

\vspace{12pt}

\end{document}